\documentclass{article}

\usepackage[preprint,nonatbib]{neurips_2023}

\usepackage[utf8]{inputenc} 
\usepackage[T1]{fontenc}    
\usepackage{hyperref}       
\usepackage{url}            
\usepackage{booktabs}       
\usepackage{amsfonts}       
\usepackage{nicefrac}       
\usepackage{microtype}      
\usepackage{xcolor}         
\usepackage{amsmath}
\usepackage{graphicx}
\usepackage{cleveref}
\usepackage{pifont}

\title{Spiking Generative Adversarial Network with Attention Scoring Decoding}

\author{%
  Linghao Feng$^{1,3}$ \qquad Dongcheng Zhao$^1$ \qquad Yi Zeng$^{1,2,3}$\thanks{Corresponding author}\\
  $^1$Brain-inspired Cognitive Intelligence Lab, Institute of Automation, Chinese Academy of Sciences\\
  $^2$Center for Excellence in Brain Science and Intelligence Technology, CAS\\
  $^3$School of Future Technology, University of Chinese Academy of Sciences\\
  \texttt{\{fenglinghao2022,\;zhaodongcheng2016,\;yi.zeng\}@ia.ac.cn}
}

\begin{document}

\maketitle

\begin{abstract}

    Generative models based on neural networks present a substantial challenge within deep learning. As it stands, such models are primarily limited to the domain of artificial neural networks. Spiking neural networks, as the third generation of neural networks, offer a closer approximation to brain-like processing due to their rich spatiotemporal dynamics. However, generative models based on spiking neural networks are not well studied. In this work, we pioneer constructing a spiking generative adversarial network capable of handling complex images. Our first task was to identify the problems of out-of-domain inconsistency and temporal inconsistency inherent in spiking generative adversarial networks. We addressed these issues by incorporating the Earth-Mover distance and an attention-based weighted decoding method, significantly enhancing the performance of our algorithm across several datasets. Experimental results reveal that our approach outperforms existing methods on the MNIST, FashionMNIST, CIFAR10, and CelebA datasets. Moreover, compared with hybrid spiking generative adversarial networks, where the discriminator is an artificial analog neural network, our methodology demonstrates closer alignment with the information processing patterns observed in the mouse.
\end{abstract}

\section{Introduction}\label{sec:intro}
Spiking Neural Networks (SNNs), which use spiking neurons as their basic computational units, operate more closely to biological neurons compared to traditional artificial neural networks (ANN)~\cite{maass1997networks}. This characteristic has garnered SNNs' widespread attention and application in deep learning and brain simulation~\cite{zeng2022braincog}. Notably, due to the use of discrete spike sequences to transmit information, SNNs are especially suitable for operation on neuromorphic hardware~\cite{2018Loihi,2015TrueNorth,2014Real,2014Neurogrid}. However, due to the non-differentiable nature of spiking neurons, many researchers have attempted to train SNNs using brain-inspired learning rules such as spike timing dependent plasticity (STDP). However, these methods have only shown good performance in relatively simple tasks, such as image recognition~\cite{zhao2020glsnn,dong2022unsupervised}. With the introduction of surrogate gradients, SNNs can be trained using the backpropagation algorithm, thereby achieving significant results in more complex tasks, such as object detection, object tracking~\cite{zhang2022spiking}, voice activity detection~\cite{martinelli2020spiking}, pre-trained language model~\cite{zhu2023spikegpt}. However, there is little research on SNNs for more complex generative tasks.

Generative models~\cite{ruthotto2021introduction} are a widely researched direction in deep learning. Unlike traditional recognition tasks\cite{lu2007survey,maulud2020review}, generating models need to precisely generate the value of each pixel, which is a considerable challenge for SNNs, which primarily use binarized spikes. In addition, the training of generating models typically faces challenges such as mode collaps\cite{bau2019seeing} and gradient vanishing\cite{hochreiter1998vanishing,Arjovsky2017WassersteinGA}, which make it difficult to guarantee the quality and diversity of generated images. The discrete and non-differentiable characteristics of SNNs make it challenging to fine-tune pixel values. The generation model based on SNNs is a significant challenge.

Within generative models, generative adversarial networks (GANs) are particularly emblematic and noteworthy. The primary distinction between GANs constructed on SNNs and those built using ANNs is that SNNs generate an image at every time step. Upon analysis, we find that the quality of images generated at distinct time steps varies, resulting in distortions in the images produced by GANs utilizing SNNs. To rectify this issue, we have designed an attention-based decoding mechanism based on spatial and temporal distribution consistency. This method dynamically assigns varying weights to different time steps, thereby producing higher quality images through a dynamic combination process.
In summary, the key contributions of this study can be outlined as follows:
\begin{itemize}
    \item Through analysis, we find that GANs built on SNNs exhibit issues of uneven sample distribution across multiple time steps, which results in the collapse and severe distortion of the generated images.
    \item We put forward an attention-based decoding mechanism that hinges on the consistency of spatial and temporal distributions, effectively assigning varying weights to different time steps dynamically.
    \item To demonstrate the superiority of our algorithm, we conduct experiments on multiple datasets, including MNIST, FashionMNIST, CIFAR10, and the CelebA dataset. The results demonstrate that our model significantly outperforms existing generative models based on SNNs.
\end{itemize}

\begin{figure}[t]
    \centering
    \includegraphics[width=1.0 \textwidth]{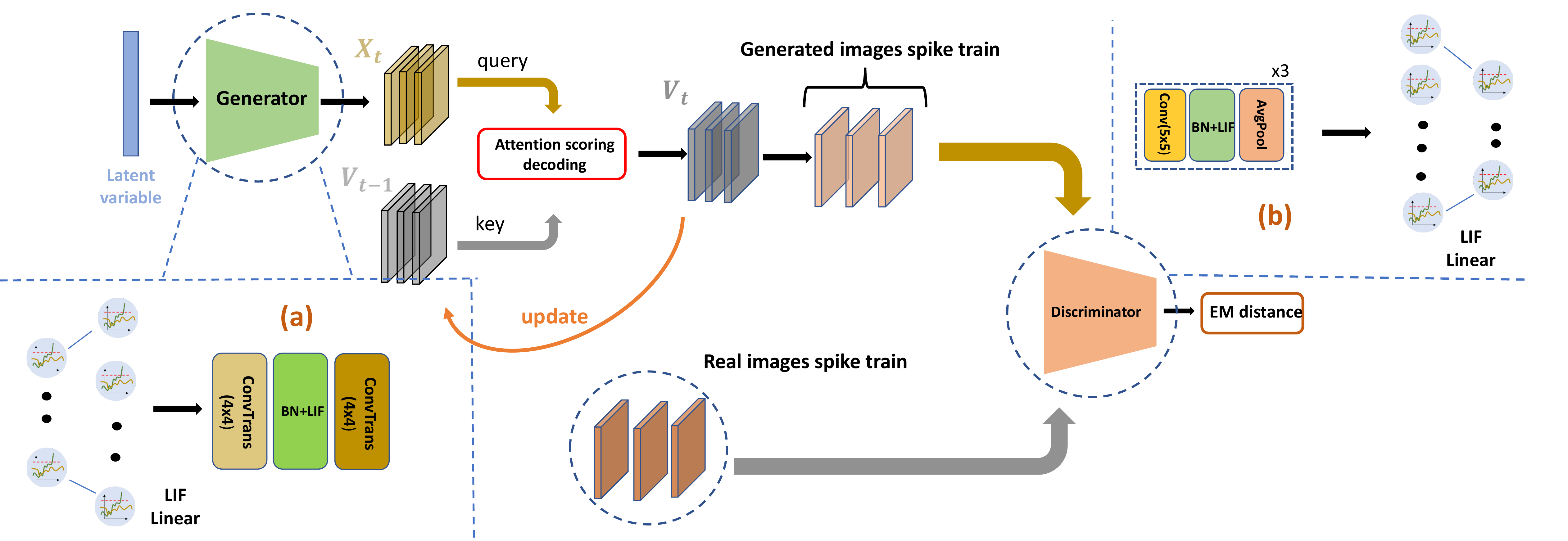}
    \caption{
        The overview of SGAD. The generator outputs membrane potential increment $X_t$. Then attention scoring decoding combine $X_t$ and $V_{t-1}$ to update $V_t$. The loss function of the discriminator is EM distance. The discriminator and generator's detailed structures are illustrated in parts (b) and part (a).}
    \label{fig:overview}
\end{figure}
\section{Related Work}\label{sec:related}

\subsection{Development of SNNs}
Currently, SNNs trained based on the backpropagation algorithm are committed to solving the problems of precise spikes firing. NeuNorm~\cite{wu2019direct}, tdBN~\cite{zheng2021going}, and TEBN~\cite{duan2022temporal} have proposed Norm-based methods to address the spike vanishing problems in SNNs, enabling SNNs to be applicable to deeper structures. ~\cite{shen2023exploiting} method has greatly enhanced the representational capability of spiking neural networks by introducing a burst-based spike firing pattern. Some researchers control the transmission of spike information more precisely through attention-based methods~\cite{yao2021temporal,zhu2022tcja,yao2023attention}. Researchers have constructed higher-performance spiking neural networks by reducing information loss by converting membrane potential to spikes~\cite{guo2022reducing,guo2022loss}. Additionally,   BackEISNN~\cite{zhao2022backeisnn} and LISNN~\cite{cheng2020lisnn} improve the performance and convergence speed of spiking neurons by improving the structure. However, these methods are improvements based on recognition tasks, and there is little work designing SNNs for more complex tasks.

\subsection{Generative Models Based on SNNs}
Following the original form of variational autoencoder~\cite{2014Auto} (VAE), researchers have proposed generative models such as nearly optimal VAE~\cite{bai2020nearly}, deep VAE~\cite{hou2017deep}, Hamiltonian VAE~\cite{caterini2018hamiltonian} and hierarchical Nouveau VAE~\cite{vahdat2020nvae}. As for VAE based on SNNs, ~\cite{kamata2022fully} utilized an autoregressive SNN model to construct the latent space, building a full spiking VAE that can generate performance comparable to traditional ANNs. Following the original form of generative adversarial network~\cite{2014Generative}, generative models like Conventional GAN\cite{sixt2018rendergan}, Wasserstein GAN\cite{Arjovsky2017WassersteinGA}, Bayesian Conditional GAN\cite{aggarwal2021generative}, DCGAN\cite{teramoto2020deep}, and SC-GAN\cite{lan2020sc}. As for GANs based on SNNs, ~\cite{kotariya2022spiking} uses the time-to-first spike coding method and builds a spiking generative adversarial network (GAN), which generates samples of higher quality compared to GANs based on ANNs. However, these models could only be validated on relatively simple datasets like MNIST. The main contribution of our work compared to ~\cite{kotariya2022spiking} is that we design a special decoding method for generative models based on SNN, and we achieve better generation quality across several datasets.

\section{Preliminary}
In this section, we will introduce the basic knowledge of the spiking neuron models and GANs that we use.
\subsection{Spiking Neuron Model}
This paper employs the Leaky Integrate-and-Fire (LIF) neuron model. The LIF model uses a differential equation as shown in Eq.~\ref{eq:lif_dy1} to describe the accumulation process of neuronal membrane potential. It strikes a good balance between biological realism and computational convenience, thus it is widely used in the modeling of deep spiking neural networks.
\begin{equation}
    \tau \frac{dV}{dt}= -V+ X
    \label{eq:lif_dy1}
\end{equation}

To facilitate computation, we provide its discrete form as shown in Eq.~\ref{o}.
\begin{align}
     & V^t = (1 - \frac{1}{\tau})V^{t-1} + \frac{1}{\tau}X^t  \label{lif2} \\
     & O^t = H(V^t-V_{th}) \label{o}
\end{align}
where $V^t$ is the membrane potential at time step $t$. And $X^t=\sum_{i} w_{j,i} O_{i}^t$ the presynaptic current. $\tau$ is the membrane potential constant. When membrane potential $V^t$ exceeds a certain threshold $V_{th}$, the neuron fires a spike $O^t$. Then, $V^t$ is reset to the reset potential $V_r$. Here, we set $V_r=0$. $H(x)$ rdenotes the step function in the spike firing process. Here, we use surrogate gradients for optimization.

\subsection{Generative Adversarial Network}
First, we briefly introduce the basic concept of GAN. GAN includes a discriminator $D$ and a generator $G$. The task of the discriminator D is to distinguish whether the input data is real or generated by the generator $G$. The goal of the generator $G$ is to generate data as similar as possible to real data, so that the discriminator D cannot easily distinguish. In other words, $D$ and $G$ play the following two-player minimax game with value function $V (G; D)$:
\begin{equation}
    \begin{split}
        &\underset{G}{min} \; \underset{D}{max} \;V(D,G) =\mathbb{E}_{x \sim p_{data}(x)}\left[\log D(x)\right] + \mathbb{E}_{x \sim p_{data}(x)}\left[\log D(x)\right]
    \end{split}\label{minmax}
\end{equation}
Here, $D(x)$ represents the probability provided by the discriminator, which measures the similarity between the input image $x$ and the data distribution. On the other hand, $G(z)$ corresponds to the image produced by the generator.

\section{Methods}
SNNs gradually receive input information over time. When we attempt to use an image as input for the SNN, static image information needs to be encoded into a spike sequence along the temporal dimension. Correspondingly, when the SNN generates output signals, we also need to decode useful information from outputs at different moments. There are various decoding methods, and the most common one currently is to use the average output of all moments in the last layer.

\begin{figure}[htbp]
    \centering
    \includegraphics[width=1.0 \textwidth]{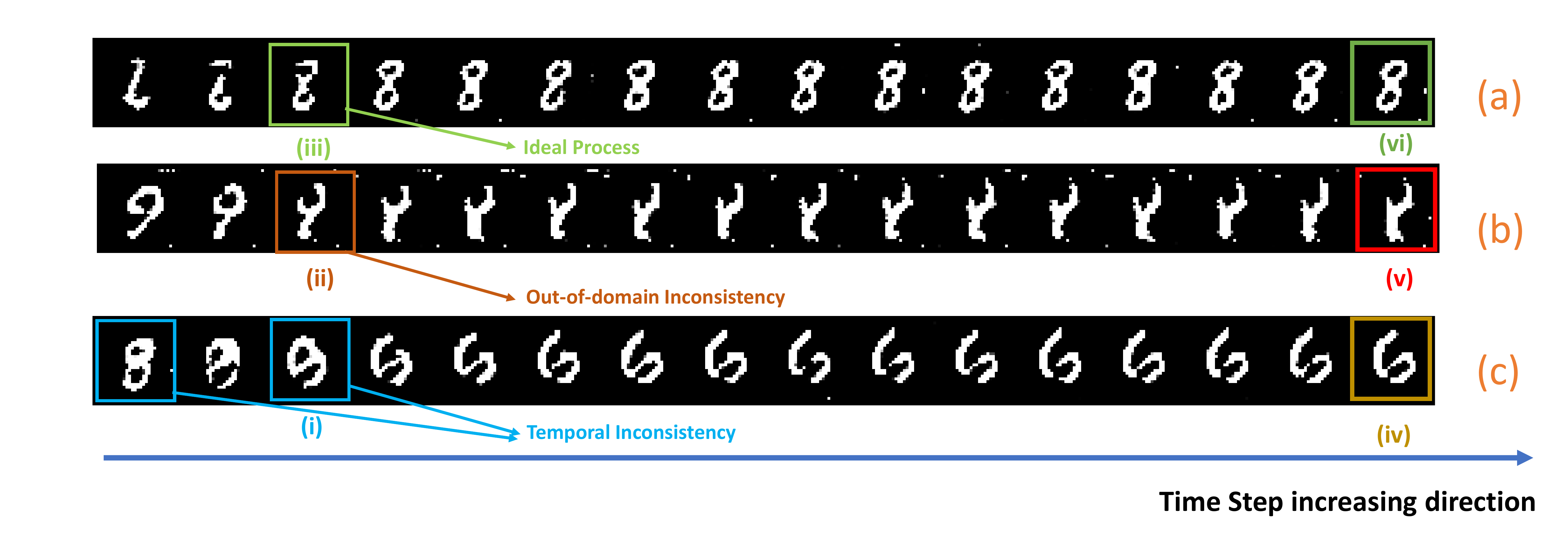}
    \caption{Process (a) is an ideal generating process. Process (b) violates out-of-domain consistency. The red box (ii) denotes a shape inconsistent with MNIST distribution. Process (c) violates temporal consistency. The blue boxes (i) denote two shapes similar to different numbers. The boxes (iv)(v)(vi) are outputs at the final time step.}
    \label{collapse}
\end{figure}

However, the information of SNN at different moments is usually inconsistent. Unlike classification problems, even if errors occur at one or two moments, it will not cause severe impacts. But for SNNs based on generative tasks, a collapsed distribution at a certain moment may cause distortion of the overall image. As shown in Fig.~\ref{collapse}, we demonstrate the situation of SNN generating images at different moments based on the MNIST dataset. Ideally, as shown in Fig.~\ref{collapse}.a, as the time step increases, the shape of the number "8" gradually becomes clear. The other two situations will lead to poor generation of images, which we call "out-of-domain inconsistency" and "temporal inconsistency" problems.

For the issue of out-of-domain inconsistency, as depicted in Fig.~\ref{collapse}.b, the model might generate a multitude of images that do not belong to the dataset, indicating a failure in adequately approximating the actual images contained within the dataset. We attribute this to the generator not converging to a suitable point. Consequently, we have employed enhanced training strategies and optimized loss functions to address this issue.

With respect to temporal inconsistency, as demonstrated in Fig.~\ref{collapse}.c, the images produced by the algorithm all align with the actual data distribution within the dataset, yet they are not consistent across different time steps. For instance, the generator may produce an image resembling the number "2" at time $T=3$, but by $T=6$, it generates an image more similar to the number "7". This discrepancy frequently arises due to the similarities between the shapes of the numbers "2" and "7". During decoding, the model must integrate these two outputs. If equal weight is assigned to both the numbers "2" and "7", the resultant image becomes blurred and lacks stability. To penalize this inconsistency, the decoding mechanism needs to assign a lesser weight to the number "7".
\subsection{Overview of Structure}
The overview of SGAD is illustrated in Fig.\ref{fig:overview} For generator part, the latent variables go through generator backbone, and the generator outputs membrane potential increment $X_t$. After that, attention scoring decoding gives $X_t$ a weight and combines it into $V_{t-1}$ to obtain $V_t$. When $t$ equals the number of total time steps, the generator gives the final images. For discriminator part, virtual images and real images are encoded and fed into the discriminator. The loss function of discriminator is EM distance. The discriminator consists of three downsample blocks. Each block comprises a convolution layer, a LIF neuron, and an average pooling layer, as illustrated in part (b) of Fig.\ref{fig:overview}. We utilize average pooling here because it can better transmit information for binary spike neurons.
The generator consists of a linear network and two transposed convolution layers, as illustrated in part (a) of Fig.\ref{fig:overview}. In the bottlenecks of discriminator and generator, we utilize a LIF neuron's membrane potential to output logits. Because membrane position can be float numbers rather than binary signals, it can better represent probability and improve the quality of complex downstream tasks.
\subsection{Out-of-Domain Inconsistency}\label{sec:out-of-domain}
In the original GAN, when the discriminator converges too fast, it may cause the generator to not obtain ideal results. This problem is particularly severe in GANs based on spiking neural networks. We analyze the reason for this issue in \cref{sec:reason}. To mitigate this issue, we borrow the  Earth-Mover distance from WGAN~\cite{Arjovsky2017WassersteinGA}. Specifically, we modify the loss function of the discriminator as shown in Eq.~\ref{wgan}:
\begin{equation}
    \underset{\Vert f \Vert \leq 1}{max} \; \mathbb{E}_{x \sim P_r}[f(x)]-\mathbb{E}_{x \sim P_\theta}[f(x)]
    \label{wgan}
\end{equation}
where $f$ denotes the mapping function of the discriminator. $P_r$ is the dataset distribution and $P_{\theta}$ is generated images distribution. $\Vert f \Vert \leq 1$ means $f$ satisfy 1-Lipschitz condition. To satisfy the 1-Lipschitz condition, we chose the RMSProp optimizer, the same as used in WGAN, but we refrain from using weight clipping. This is because when the weights are small, weight clipping could lead to the inability to properly activate the deeper neurons, resulting in the "spike vanishing" problem, which in turn prevents the Spiking GAN from generating images properly.
\subsection{Temporal Inconsistency}
Regarding temporal consistency, we contend that the final image decoded from the SNN can be considered a function of images generated at each time step. Our goal is not only to discard images of poor quality while retaining those of high quality but also to ensure the temporal consistency of the images generated by the SNN. For instance, if images generated at different time steps correspond to different shape labels - such as an image at $T=1$ resembling the number "1" and an image at $T=3$ resembling the number "7" - this discrepancy could degrade the quality of the final result.

As shown in Eq.~\ref{lif2}, the membrane potential at time $t$ is determined by the membrane potential at the previous time step $V_{t-1}$ and the presynaptic input $X_t$. We can dynamically adjust $\tau$ to balance the output at the current moment with the information from the historical moment and the input information at the current moment.
\begin{figure}[htbp]
    \centering
    \includegraphics[width=0.6 \linewidth]{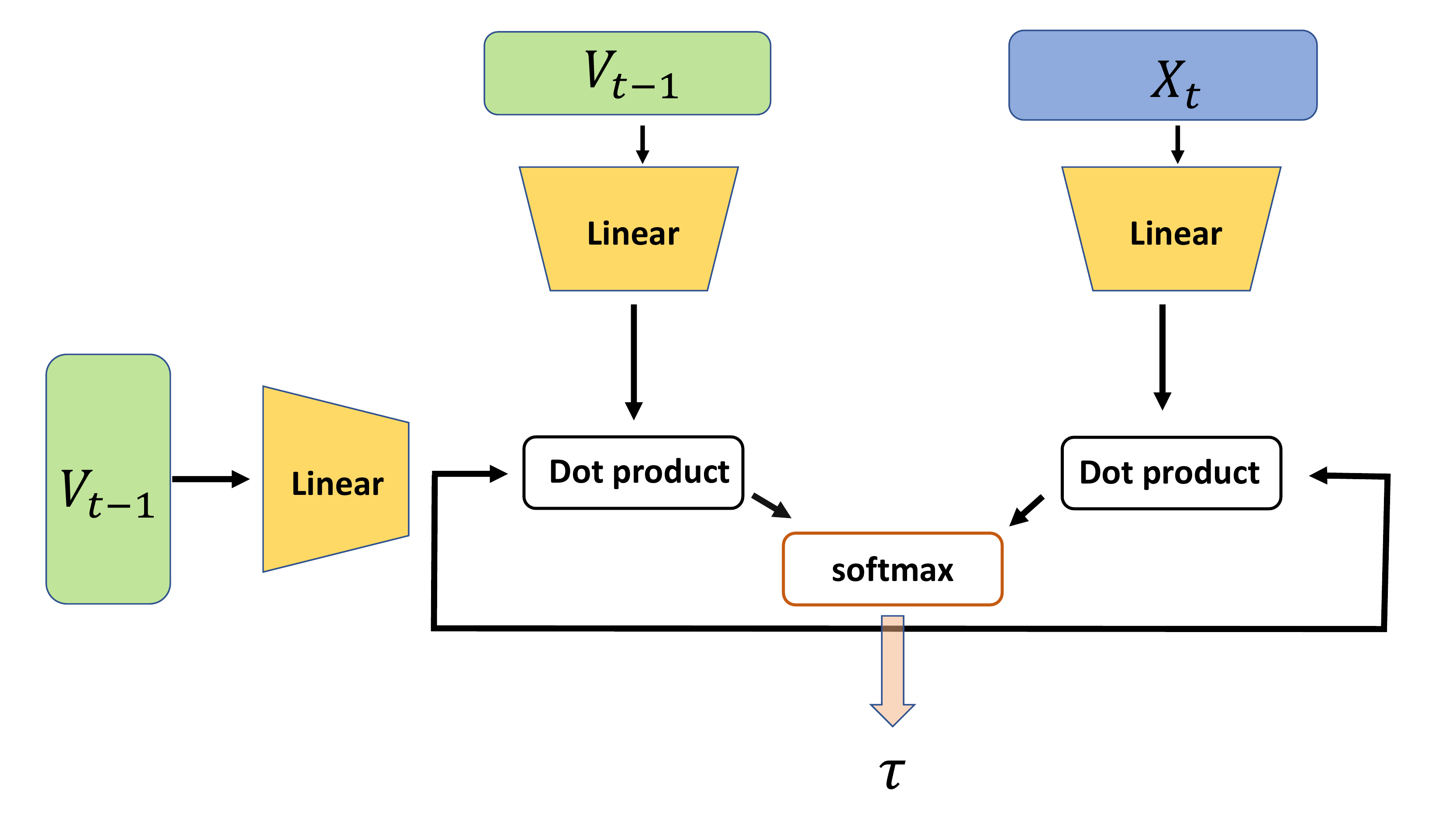}
    \caption{The sketch of attention scoring decoding method.}
    \label{fig:attention_decoding}
\end{figure}

To this end, we use an attention scoring function to calculate the appropriate value of $\tau$. The schematic of this attention scoring function decoding method is shown in Fig.~\ref{fig:attention_decoding}. In this setup, the membrane potential at time $t-1$ ($V_{t-1}$) and the pre-synaptic input at time $t$ ($X_t$) are critical to the attention scoring function, with the membrane potential at $t-1$ ($V_{t-1}$) acting as the query and pre-synaptic input $X_t$ together with $V_{t-1}$ itself as key. Then, the query and the key are passed through a dense network to obtain latent variables. Finally, through scaled dot product and softmax, we get two probabilities. These two probabilities are assigned to the values of $1-\frac{1}{\tau}$ and $\frac{1}{\tau}$, respectively. The above process can be expressed as the following formula:
\begin{equation}
    [\frac{1}{\tau}, 1-\frac{1}{\tau}] = AttentionScore(V_{t-1},X_t)=softmax(\frac{f_{v_q}(V_{t-1}){[f_{x_k}(X_t), f_{v_k}(V_{t-1})]}^T}{\sqrt{d_k}})
\end{equation}
where $\frac{1}{\sqrt{d_k}}$ is scaling factor. $f_{v_q}$ is the linear mapping for the query $V_{t-1}$, and $f_{v_k}, f_{x_k}$ are linear mapping for keys $V_{t-1}, X_{t}$. We analyze the validity of above method in ~\cref{sec:validity}

\subsection{Hybrid GAN} \label{sec:Hybrid GAN}
We also built a hybrid GAN to compare with GAN purely based on SNN. Its discriminator utilizes ANN, and the generator utilizes SNN. The idea of combining SNN and ANN to build a generative adversarial network has been proposed before\cite{9832788}. We change its structure and test it on several datasets. Hybrid GAN is not the primary research purpose of this paper. We build it to compare with other models and explore insights of GAN purely based on SNN.

\section{Experiments}\label{sec:exp}
To validate our algorithm, we conduct experiments on multiple datasets, including MNIST, Fashion MNIST~\cite{2017Fashion}, CIFAR10, and CelebA~\cite{liu2015faceattributes}. For MNIST, Fashion MNIST, and CIFAR10, we keep the original image size. For the CelebA dataset, we resized the images to a resolution of 64$\times$64. We use Fréchet Inception Distance (FID) ~\cite{2017GANs,2021On} score to evaluate the generated image quality of four datasets. For all datasets, we train SGAD to 200 epochs, and the total time step is $T=16$. For the first 100 epochs, we use a constant learning rate of discriminator and generator. Moreover, the following 100 epochs, we use CosineAnnealing scheduler to adjust learning rate.  For different datasets, we use different latent dimensions for the generator. The latent dimension of MNIST is 10. And for CIFAR10 and FashionMNIST, the latent dimensions are both 15. For CelebA, the latent deminsion is 20.
\subsection{The Improvement of Gradient Stability}\label{sec:reason}
As mentioned in \cref{sec:out-of-domain}, the GAN based on SNN with minimax as loss function has out-of-domain inconsistency problem. So we take the idea of WGAN, change the loss function to EM distance, and change its optimizer. According to the theory of WGAN\cite{Arjovsky2017WassersteinGA}, minimax loss will cause gradient vanishing. Furthermore, in GAN based on SNN, we think this issue will even be worse due to the inaccurate approximation of surrogate gradient and ``spike vanishing'' problem of SNN. In order to analyze this problem, we output the gradient of different models.

\begin{figure}[htbp]
    \centering
    \includegraphics[width=0.6 \linewidth]{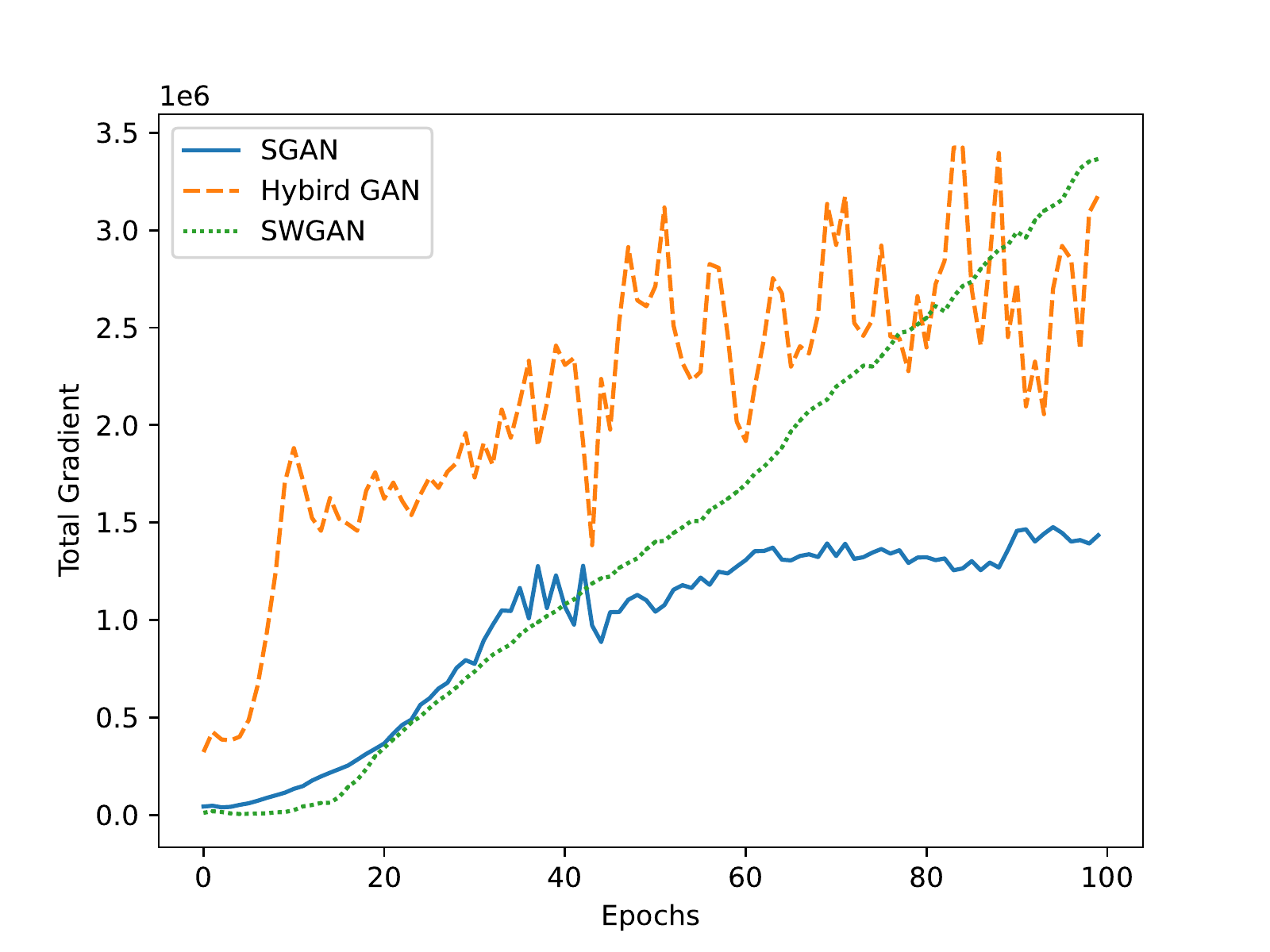}
    \caption{The gradients of different models vary with epochs. We could conclude that 1. SGAN suffers gradient vanishing problem compared to the other two models. 2. The gradient vanishing problem is more significant for the model based on SNNs than the model based on ANNs}
    \label{fig:grads}
\end{figure}

Fig.~\ref{fig:grads} shows the total gradient of each epoch during a training process of three different models: SGAN, SWGAN, and Hybrid GAN. SGAN is the GAN based on SNN with minimax loss. SWGAN is the GAN based on SNN with EM distance loss.
And Hybrid GAN is illustrated in \cref{sec:Hybrid GAN}.
We could see that the gradient of Hybrid GAN is generally higher than others. The gradients of SWGAN and SGAN are similar initially. However, after epoch 40, the gradient of SWGAN is higher than SGAN and gradually similar to Hybrid GAN. It could be explained that the vanishing of gradient causes the worse performance of SGAN.
\subsection{Validity of Attention Scoring Decoding Method}\label{sec:validity}
\begin{figure}[htbp]
    \centering
    \includegraphics[width=1.0 \linewidth]{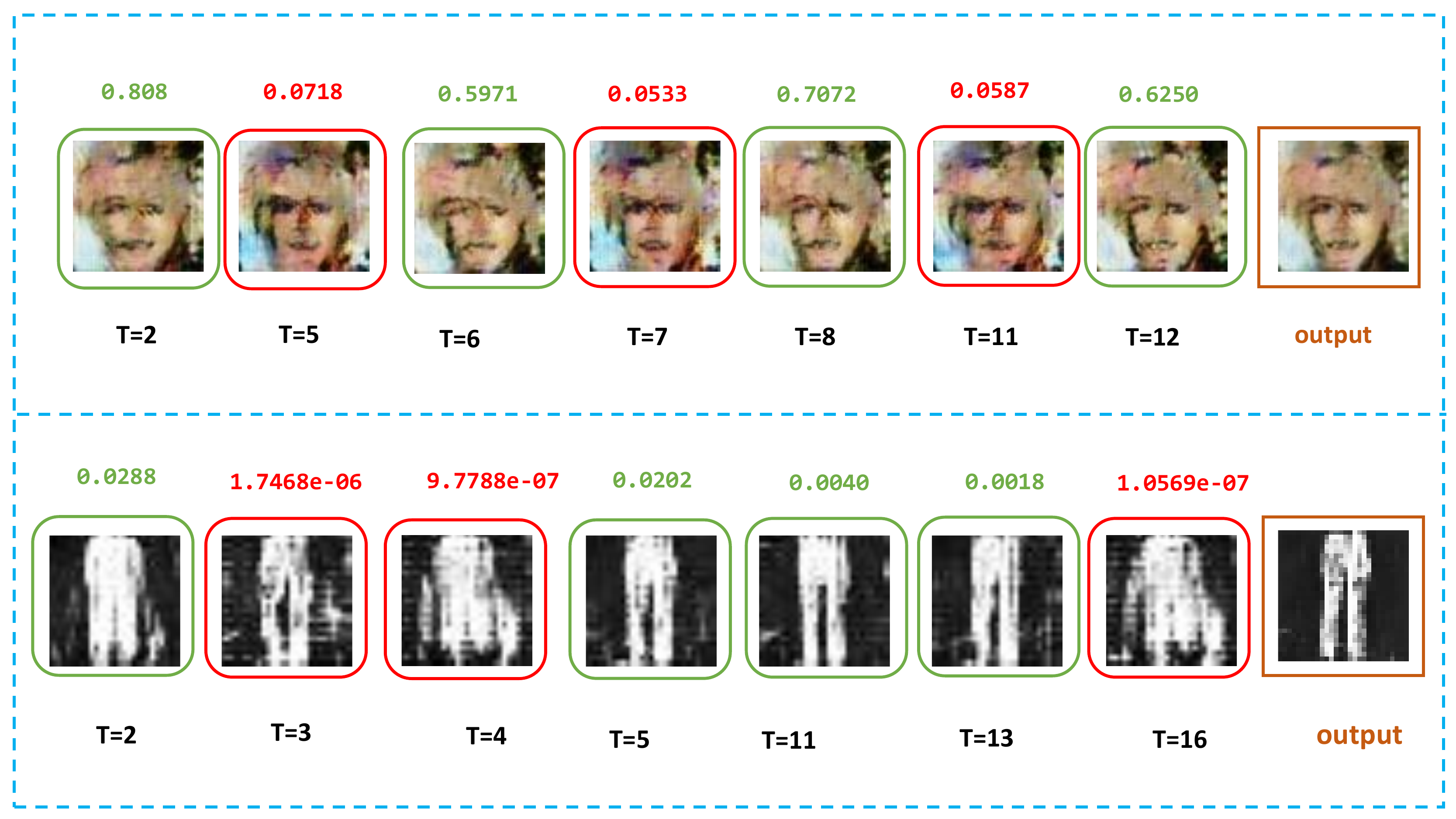}
    \caption{The presynaptic($X_t$) scores given by attention scoring decoding method. The red boxes denote presynaptic inputs with lower scores, while the green boxes denote presynaptic inputs with higher scores. The above line is the generating process of CelebA, and the bottom line is the process of FashionMNIST. We could see that those presynaptic inputs with distortion or darker and sharper colors are given lower scores.}
    \label{fig:score}
\end{figure}

To verify the validity of our method, we show two specific images generating process. One generating process is from Fashion MNIST and the other is from CelebA. As is shown in Fig.\ref{fig:score}, we print the presynaptic inputs ($X_t$ in Eq.\ref{lif2}) into the bottleneck neurons and give their scores ($\frac{1}{\tau}$ in Eq.\ref{lif2}) judged by attention scoring decoding method.

From Fig.\ref{fig:score}, we could first see that attention scoring decoding method gives higher scores for those presynaptic inputs similar to final outputs. For example, from the generating process of human face (above line), attention scoring decoding method selects presynaptic inputs with darker and sharper color($T=5,7,11$) compared with the final output. It gives their lower scores, although these presynaptic inputs have no noticeable distortion compared with the final output. It could be explained that the attention scoring decoding method first selects a statistical center and filters those presynaptic inputs deviating from this statistical center, which could contribute to ``time step consistency''. Also, attention scoring decoding method can filter those presynaptic inputs that have a distortion in shape. The bottom line of Fig.\ref{fig:score} shows a generating process of FashionMNIST. From it we could see that attention scoring decoding method tends to give presynaptic inputs with distortion($T=3,4,16$) lower scores, which could contribute to ``distribution consistency''.

From Fig.\ref{fig:score}, it is not hard to find that the value of scores given by attention scoring decoding have different orders of magnitude. The order of magnitude of presynaptic input scores for CelebA is ${10}^{-1} \sim {10}^{-2}$. However, for FashionMNIST, it is ${10}^{-2} \sim {10}^{-7}$.

\subsection{Ablation Study}
To compare different models mentioned in this paper, we list four different models built in this paper, as shown in \cref{tab:res}. SGAN is the original GAN based on SNN, which utilizes minimax loss. SWGAN mainly changes the loss function of SGAN to EM distance. SGAD is added attention scoring decoding method on the basis of SWGAN. Hybrid GAN is a combination of SNN and ANN. Its discriminator utilizes ANN, and generator utilizes SNN, as mentioned in \cref{sec:Hybrid GAN}

We compare the performance of 4 pure SNN models in 4 datasets: MNIST, FashionMNIST, CIFAR10, and CelebA. SGAD improves image quality in three datasets. However, SGAD does not perform better than the model without attention in CIFAR10. We analyzed it and found that the generated images based on the model without attention on CIFAR10 are initially bad, so attention scoring decoding may fail to recognize the quality of signals and cannot improve the result.

\begin{figure}[t]
    \centering
    \includegraphics[width=1.0 \linewidth]{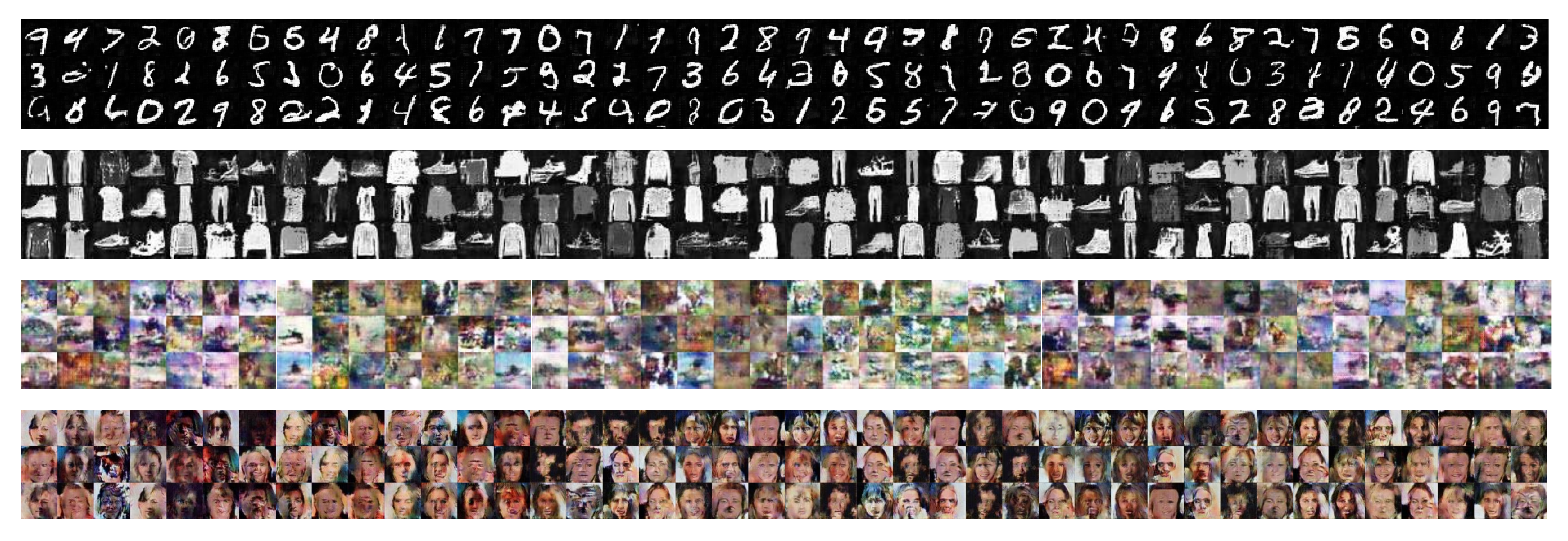}
    \caption{The results of SGAD on MNIST, FashionMNIST, CIFAR10, and CelebA.}
    \label{fig:res}
\end{figure}

We also test the performance of Hybrid GAN. Intuitively speaking, Hybrid GAN should perform better than GAN purely based on SNN, because the performance of ANN is generally better than SNN. The result shows Hybrid GAN outperforms the other four models on complex datasets (CIFAR10, CelebA). For simple datasets (MNIST, FashionMNIST), SGAD outperforms Hybrid GAN, which means attention scoring decoding method could enhance SNN to overcome ANN to some degree. The reason SGAD performs worse than Hybrid GAN on complex datasets is that SNNs are harder to handle deeper network structures than ANN. Some outputs of SGAD is shown in \cref{fig:res}

\begin{table}[t]
    \centering
    \begin{tabular}{c|ccc|c}
        \hline
        Datasets/Methods               & SGAN      & SWGAN           & SGAD            & Hybrid GAN             \\
        \hline
        Loss                           & minimax   & EM distance     & EM distance     & minimax                \\
        \hline
        Discriminator                  & SNN       & SNN             & SNN             & \textbf{ANN}           \\
        \hline
        Generator                      & SNN       & SNN             & SNN             & SNN                    \\
        \hline
        Use attention scoring decoding & \ding{55} & \ding{55}       & \ding{51}       & \ding{55}              \\
        \hline
        FID on MNIST                   & 203.28    & 100.29          & \textbf{69.64}  & 123.93                 \\
        \hline
        FID on FashionMNIST            & 176.57    & 175.34          & \textbf{165.42} & 198.94                 \\
        \hline
        FID on CIFAR10                 & ---       & \textbf{178.40} & 181.50          & \textcolor{red}{72.64} \\
        \hline
        FID on CelebA                  & ---       & 238.42          & \textbf{151.36} & \textcolor{red}{63.18} \\
        \hline
    \end{tabular}
    \caption{The Fréchet Inception Distance (FID) score of models mentioned in this paper on four datasets. Compared to pure SNN models, SGAD improves image quality in most datasets. For simple datasets (MNIST, FashionMNIST), SGAD outperforms Hybrid GAN.}
    \label{tab:res}
\end{table}

\subsection{Biological Similarity Analysis}
\begin{figure}[t]
    \centering
    \includegraphics[width=1.0 \linewidth]{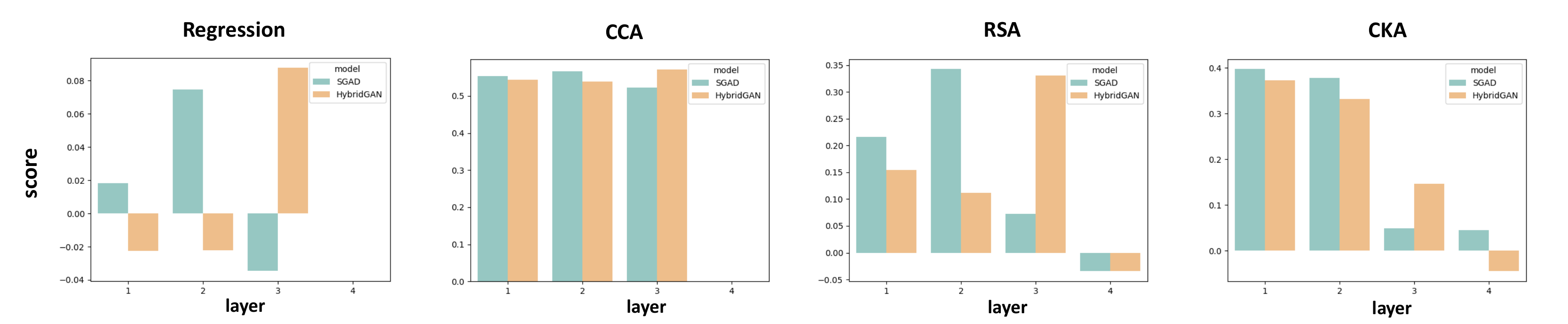}
    \caption{Comparisons of similarity scores of SGAD and HybridGAN with RSA, Regression-Based Encoding Method, SVCCA, and CKA in a specific brain area visp}
    \label{fig:metrics}
\end{figure}

\begin{figure}[t]
    \centering
    \includegraphics[width=1.0 \linewidth]{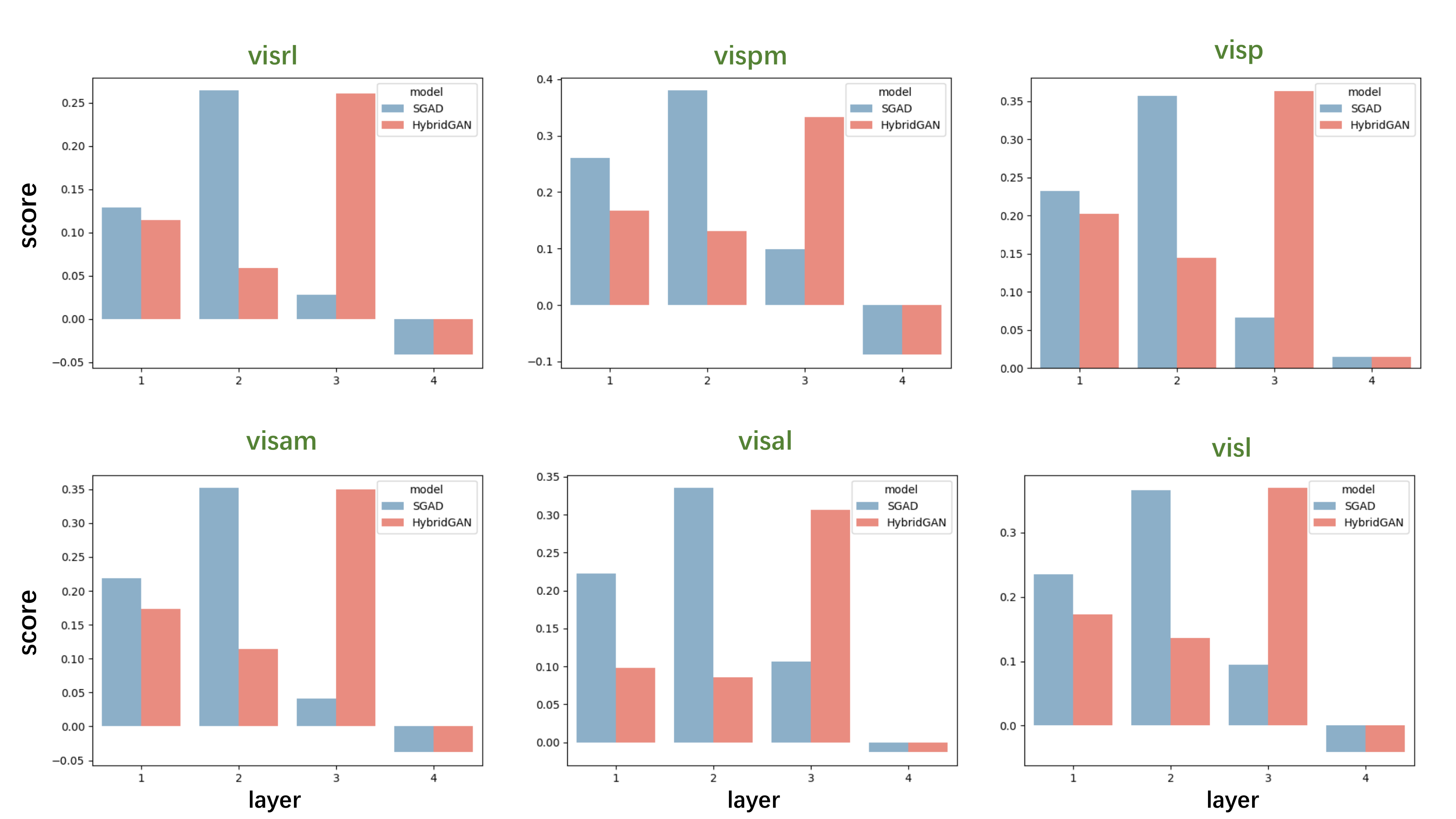}
    \caption{Comparisons of RSA scores of SGAD and HybridGAN through six brain areas: primary (visp), lateral (visl), anterolateral (visal), anteromedial (visam), posteromedial (vispm), and rostrolateral (visrl) visual areas}
    \label{fig:areas}
\end{figure}

We compare the biological similarity of real mouse brain spiking series with SGAD and HybridGAN proposed in this paper. Firstly, we extracted the stimulus in the discriminators of SGAD and HybridGAN layer by layer. Then we applied four different similarity metrics to measure the similarity for different layers between SGAD, HybridGAN, and real biological mouse brains. The mouse brain stimulus data is from Allen Brain Observatory Visual Coding dataset\cite{siegle2021survey} collected using Neuropixel probes from 6 regions simultaneously in mouse visual cortex. The four similarity metrics are RSA\cite{kriegeskorte2008representational,kriegeskorte2008matching}, Regression-Based Encoding Method\cite{carandini2005we,yamins2014performance,schrimpf2018brain,schrimpf2020integrative}, SVCCA\cite{raghu2017svcca,morcos2018insights} and CKA\cite{hardoon2004canonical}. We compare the scores with different metrics in primary visual areas (visp), as is shown in \cref{fig:metrics}. We show the similarity scores through six different brain areas with the RSA metrics, as is shown in \cref{fig:areas}. Through \cref{fig:areas} and \cref{fig:metrics}, we could conclude that the similarity scores of SGAD for shallow layers are higher than HybridGAN. This means SGAD is more biologically plausible than HybridGAN for shallow layers, which also indicates discriminator built by SNN is more biologically plausible than ANN.

\section{Conclusion}
In this paper, we first explore the issues prevalent in generative adversarial networks based on spiking neural networks. We found that the primary issues stem from out-of-domain inconsistency and temporal inconsistency, i.e., SNNs might generate images that are inconsistent with the sample distribution in the dataset, and there can be substantial differences between the samples generated at different time steps. To address these issues, we introduced the Earth-Mover distance to improve the learning process of the discriminator and employed an attention-based strategy to dynamically assign different weights to each moment. Experimental results demonstrate that our approach is the first SNN-based GAN capable of handling complex datasets, exhibiting superior FID values across multiple datasets. Moreover, by comparing with the spiking signals of mouse, we find that the SNN-based discriminator exhibits a higher level of biological plasticity. This finding further confirms the excellent performance and potential of GANs based on SNNs.

\section*{Acknowledgement}
This work was supported by the National Key Research and Development Program (Grant No. 2020AAA0104305), and the Strategic Priority Research Program of the Chinese Academy of Sciences (Grant No. XDB32070100).

    {\small
    \bibliographystyle{unsrt}
    \bibliography{ref.bib}
    }

\end{document}